\documentclass[conference]{IEEEtran}
\usepackage{multirow}
\IEEEoverridecommandlockouts

\usepackage[letterpaper,left=0.64in,right=0.64in,top=0.75in,bottom=1.013in]{geometry}
\usepackage{cite}
\usepackage{amsmath,amssymb,amsfonts}
\usepackage{algorithmic}
\usepackage{graphicx}
\usepackage{textcomp}
\usepackage{xcolor}
\def\BibTeX{{\rm B\kern-.05em{\sc i\kern-.025em b}\kern-.08em
    T\kern-.1667em\lower.7ex\hbox{E}\kern-.125emX}}
\begin{document}

\title{Feature-Weighted MMD-CORAL for Domain Adaptation in Power Transformer Fault Diagnosis}

\author{
\begin{tabular}{@{\hspace{0.5cm}}c@{\hspace{2cm}}c}
\begin{tabular}{c}
\textbf{Hootan Mahmoodiyan} \\
Faculty of Engineering and Computer Science \\
University of Victoria, Canada \\
hootanmahmoodiyan@uvic.ca
\end{tabular}
&
\begin{tabular}{c}
\textbf{Maryam Ahang} \\
Faculty of Engineering and Computer Science \\
University of Victoria, Canada \\
maryamahang@uvic.ca
\end{tabular}
\\
\\
\begin{tabular}{c}
\textbf{Mostafa Abbasi} \\
Faculty of Engineering and Computer Science \\
University of Victoria, Canada \\
abbasi@uvic.ca
\end{tabular}
&
\begin{tabular}{c}
\textbf{Homayoun Najjaran} \\
Faculty of Engineering and Computer Science \\
University of Victoria, Canada \\
najjaran@uvic.ca
\end{tabular}
\end{tabular}
}

\maketitle

\begin{abstract}
Ensuring the reliable operation of power transformers is critical to grid stability. Dissolved Gas Analysis (DGA) is widely used for fault diagnosis, but traditional methods rely on heuristic rules, which may lead to inconsistent results. Machine learning (ML)-based approaches have improved diagnostic accuracy; however, power transformers operate under varying conditions, and differences in transformer type, environmental factors, and operational settings create distribution shifts in diagnostic data. Consequently, direct model transfer between transformers often fails, and therefore, techniques for domain adaptation become a necessity. To tackle this issue, in this work, a feature-weighted domain adaptation technique is proposed, combining Maximum Mean Discrepancy (MMD) and Correlation Alignment (CORAL) with feature-specific weighting (MCW). Kolmogorov–Smirnov (K-S) statistics are utilized for assigning adaptable weights, prioritizing features with larger distributional discrepancies and therefore allowing for a better source and target domain alignment. Experimental evaluations over datasets for power transformers reveal the effectiveness of our proposed method, which attains a 7.9\% improvement over Fine-Tuning and a 2.2\% improvement over MMD-CORAL (MC). In addition, it outperforms both techniques under various training sample sizes, confirming its reliability in domain adaptation.
\end{abstract}


\section{Introduction}
Power transformers are the building blocks of electric power grids, and their safe operation is crucial to grid stability. A fault in a power transformer can cause power outage and substantial economic losses. Fault diagnosis, therefore, is critical in fault prevention and maintaining uninterrupted operation. When a fault occurs in a transformer, some gases such as carbon dioxide (CO$_2$), methane (CH$_4$), ethane (C$_2$H$_6$), carbon monoxide (CO), hydrogen (H$_2$), oxygen (O$_2$), and nitrogen (N$_2$) are produced. Of these gases, except for CO$_2$, N$_2$, and O$_2$, the others are soluble in the transformer oil. These gases are quantified in the dielectric oil in parts per million (ppm) \cite{182866, IEC60599}. Therefore, the DGA method can be used for fault diagnosis.

The conventional DGA methods, such as the IEEE ratio, Key gas, Roger's ratio approach, and Duval triangle approach, are extensively applied in the condition monitoring of power transformers. These methods, however, are based on pre-established guidelines and experience, which may result in inconsistent and inaccurate diagnoses \cite{Nanfak2024}. Recent developments have been aimed at enhancing the interpretation of DGA through the application of artificial intelligence (AI)-based techniques to bypass these limitations. ML and deep learning (DL) have shown enhanced diagnostic accuracy by identifying patterns in DGA data automatically.

Considerable advancements have been achieved in using ML and DL for transformer fault diagnosis \cite{mogos2023data}. Classical methods (\textit{e.g.,} Support Vector Machines (SVMs) and Artificial Neural Networks (ANNs)), were tuned using bio-inspired algorithms like the Krill Herd Algorithm and Genetic Algorithms for improved classification accuracy \cite{Yang2019, Zhang2019}. More recently, DL models (\textit{e.g.,} Convolutional Neural Networks (CNNs) and Graph Convolutional Networks (GCNs)), were utilized to learn complex hierarchical features from the DGA data and thus outperform traditional approaches \cite{liao2020fault}.

Even with these advancements, one of the fundamental challenges in power transformer fault diagnosis is the shift in data distribution. For transformers, there are varying sample distributions based on variations in transformer types, environmental, and measurement conditions. Conventional ML-based models assume that training and test data have similar distributions, resulting in poor performance when used on unseen datasets \cite{laayati2023smart}. Handling these domain discrepancies is essential to achieve trustworthy fault diagnosis among various transformer fleets.

To alleviate this problem, domain adaptation techniques have been suggested to match feature distributions between source and target datasets. MMD and CORAL are two popular techniques employed to reduce distributional differences by aligning first-order and second-order statistical features, respectively \cite{liu2022deep}.
These techniques were applied to reduce distributional differences to obtain better classification performance \cite{liu2022deep}. In the same manner, hybrid models that integrate reinforcement learning with adaptive consistency regularization have shown effective performance in learning domain-invariant features \cite{mao2022fault}.
To improve accuracy and generalization, \cite{xie2023fault} proposed an ensemble transfer convolutional neural network (ETCNN) with an additional domain adaptation loss to address distribution gaps between the source and target domains. This method significantly improved generalization and demonstrated higher accuracy in fault classification in experimental tests.
Another study \cite{lin2022transfer} addressed the issue of data insufficiency in transformer health estimation. The proposed model employed Fine-Tuning and optimization techniques to transfer knowledge from a source domain to a target domain with limited labeled data and exhibited enhanced robustness and diagnosis efficiency.
These studies emphasize the need for domain adaptation in addressing the inconsistency between various transformer datasets. However, existing methods treat all features equally, overlooking and disregarding their unique contributions to domain disparity.

In this paper, we introduce a feature-weighted domain adaptation approach to improve transformer fault diagnosis by combining MMD-CORAL \cite{liu2022deep} with feature-specific weighting. By utilizing the K-S statistic, we estimate the extent of distributional shift for every feature and employ adaptive weights accordingly. This keeps the domain adaptation process focused on aligning features with the most significant disparities, thereby prioritizing critical areas of divergence between the source and target domains, thus resulting in better generalization across other transformer datasets. Our method is based on the latest domain adaptation techniques with a feature-weighted alignment module added, which is especially useful for practical applications where transformers are subjected to different conditions.

The rest of this paper is organized as follows: Section II explains the suggested methodology, data preprocessing, domain adaptation techniques, and model deployment. Section III presents the experimental design and results, with evidence of the effectiveness of the suggested method. Finally, section IV concludes the study.

\section{Methodology}
 
\subsection{Data Preprocessing}
Initially, hybrid DGA ratios were computed from the raw DGA data by integrating the logarithmic transformations of Roger's four ratios and the Percentage ratios. These ratios were found to increase the fault detection accuracy in CNN models \cite{taha2021power}.

Then the hybrid DGA ratios were transformed into two-dimensional (2D) images using the Gramian Angular Field (GAF) method, which encodes feature-level dependencies in structured data as 2D images \cite{liu2024fault}.

\subsection{Data Augmentation}

\begin{figure}[tb] 
\centering 
\includegraphics[width=0.95\columnwidth]{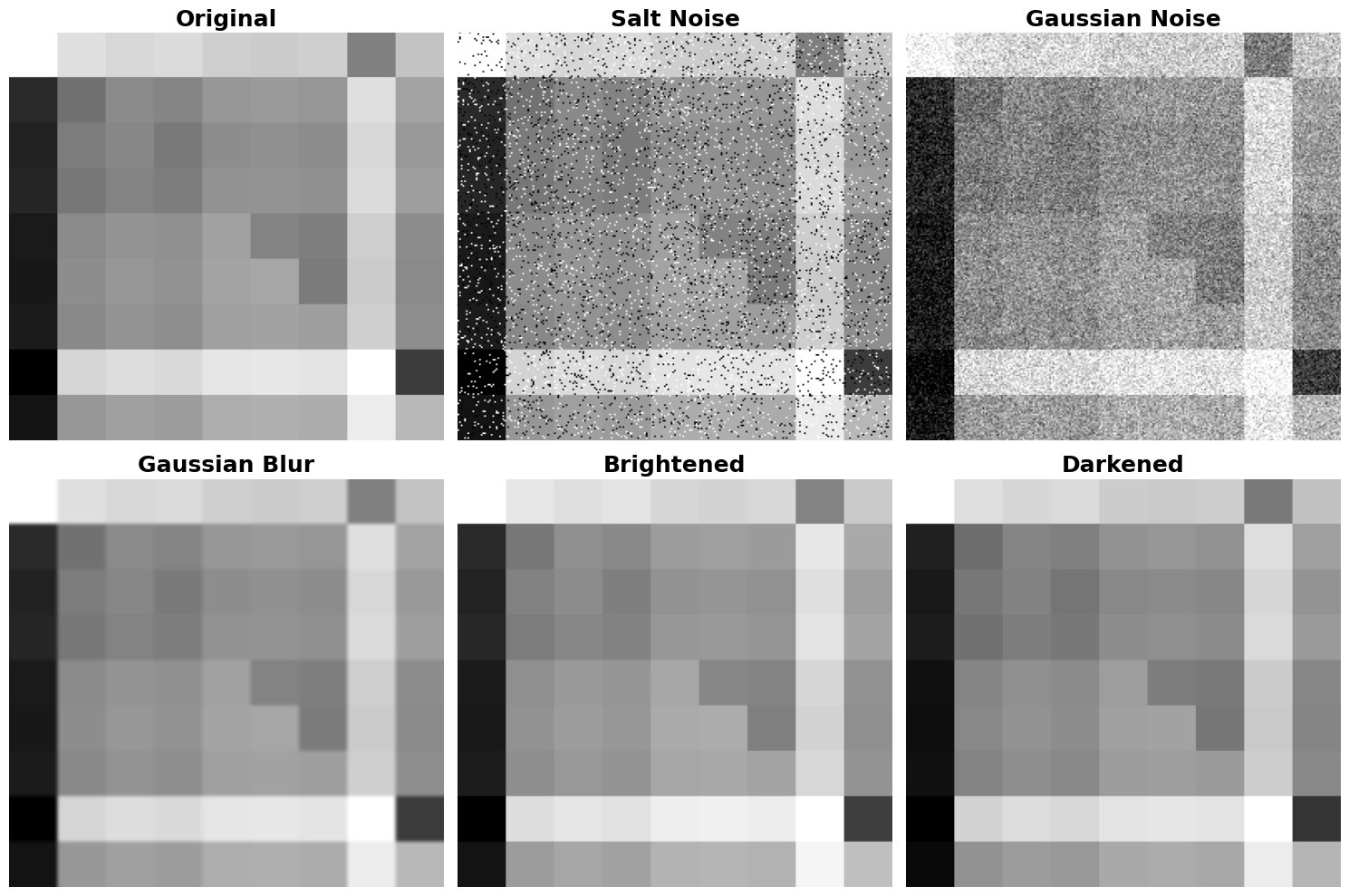} 
\caption{Augmented Images} 
\label{fig_augmentation} 
\end{figure}

Machine learning methods often struggle to deal with imbalanced data \cite{ahang2022synthesizing}. To address this issue of imbalanced label distribution and to increase the number of samples in the source dataset, we applied data augmentation techniques to the 2D Gramian Angular Field (GAF) images. The methods used governed changes in the dataset while preserving the inherent properties of the original images, thus allowing for better generalization in machine learning models. There were five methods of augmentation used: Gaussian blur, Gaussian noise, salt-and-pepper noise, brightening, and darkening \cite{liu2024fault}. An original image from the source dataset along with its augmented images is shown in Fig.~\ref{fig_augmentation}. This targeted augmentation strategy ensured a more balanced dataset by providing underrepresented classes with additional diversity and representation while preserving class-specific characteristics.

\subsection{Model Architecture}

A CNN-based model was used within this study to determine types of transformer faults from enhanced Gramian Angular Field (GAF) images. CNNs are especially suited for this kind of task as they extract hierarchical spatial features through convolution operations, enabling the model to detect subtle patterns that are intrinsic to 2D data representations \cite{liu2024fault}.

The model has two primary components which are the feature extractor and the classifier.

The feature extractor takes advantage of the strength of convolutional layers to learn spatial hierarchies by convolving filters over the input image. Mathematically, a convolutional layer takes a kernel \(K\) and an input image \(I\) to produce feature maps \(F\), as follows:
\begin{equation}
F_{i,j} = \sum_{m=-k}^{k} \sum_{n=-k}^{k} I_{i+m, j+n} \cdot K_{m,n} + b, \label{eq:conv_layer},
\end{equation}
where \(b\) is the bias, and \(k\) is the kernel size. After this operation, the ReLU activation function is applied, which adds non-linearity and enables the model to learn complex patterns. Pooling layers decrease the spatial dimensions of feature maps while maintaining the most significant features. Max pooling, utilized here, chooses the maximum value within a pool window. The classifier takes the extracted features and reshapes them as class probabilities using a fully connected (FC) network. In the classifier, the vector is passed to a dense layer, and the output vector is propagated through the ReLU activation. Dropout regularization is applied to avoid overfitting by randomly dropping a proportion of the neurons' outputs to zero.

The CNN architecture is computationally efficient. The convolutional and pooling layers progressively diminish the spatial dimension, reducing the number of parameters in the coming layers. This enables the network to concentrate on the most vital features and refrain from computational overhead.

\subsection{Weighted Domain Adaptation using K-S Statistics}

In this study, the K-S statistic output was used to enhance the effectiveness of domain adaptation by applying feature-specific weights. The weights choose features that have greater differences between the domains, based on the sizes of their K-S statistic value, thus providing selective alignment of the source and target feature distributions.

The individual feature's K-S statistics were normalized and converted into a Gramian Angular Field (GAF) image, as outlined earlier in the data preprocessing subsection. This produced a 2D weight matrix of normalized values in the range \([0, 1]\) and incorporated a small \(\epsilon > 0\) to prevent zero weights:

\begin{equation}
\text{Weights}(i, j) = \frac{\text{GAF}_{\text{weight}}(i, j) - \min(\text{GAF}_{\text{weight}})}{\max(\text{GAF}_{\text{weight}}) - \min(\text{GAF}_{\text{weight}})} + \epsilon.
\end{equation}

This weight matrix, \( \mathbf{W} \), is a 2D matrix where each entry \( W_{ij} \) indicates how significant it is to match the feature pair \(i\) and \(j\), and larger weights indicate larger domain shifts.

This weight matrix was used in MMD \cite{ghifary2014domain} and CORAL \cite{sun2016deep} methods to highlight larger domain discrepancies to be matched by the features:

The traditional computation of MMD was refined to include the weights. The weighted MMD loss is expressed as follows:  
\begin{equation}  
\begin{aligned}  
\text{MMD}^2 = & \frac{1}{n^2} \sum_{i=1}^n \sum_{j=1}^n W_{ij} k(\mathbf{x}_s^i, \mathbf{x}_s^j) \\  
& + \frac{1}{m^2} \sum_{i=1}^m \sum_{j=1}^m W_{ij} k(\mathbf{x}_t^i, \mathbf{x}_t^j) \\  
& - \frac{2}{nm} \sum_{i=1}^n \sum_{j=1}^m W_{ij} k(\mathbf{x}_s^i, \mathbf{x}_t^j),  
\end{aligned}  
\end{equation}  
where \(W_{ij}\) is the weight matrix, and \(k(\mathbf{x}, \mathbf{y})\) is the Gaussian kernel.

In addition, the weight matrix was also utilized to align the covariance in CORAL. The target and source feature covariance matrices were computed as follows:
   \begin{equation}
   \mathbf{C}_s = \frac{1}{n-1} \mathbf{W} \odot (\mathbf{X}_s - \bar{\mathbf{X}}_s)^\top (\mathbf{X}_s - \bar{\mathbf{X}}_s),
   \end{equation}
   \begin{equation}
   \mathbf{C}_t = \frac{1}{m-1} \mathbf{W} \odot (\mathbf{X}_t - \bar{\mathbf{X}}_t)^\top (\mathbf{X}_t - \bar{\mathbf{X}}_t),
   \end{equation}
   where \(\mathbf{W}\) is the weight matrix derived from the K-S statistics, \(\bar{\mathbf{X}}_s\) and \(\bar{\mathbf{X}}_t\) are the means of the source and target features, \(\mathbf{X}_s\) and \(\mathbf{X}_t\) are the feature matrices, and \(\odot\) denotes the element-wise multiplication. The weighted CORAL loss is then given by:
   \begin{equation}
   \mathcal{L}_{\text{CORAL}} = \left\| \mathbf{C}_s - \mathbf{C}_t \right\|_F^2,
   \end{equation}
   where \(\|\cdot\|_F^2\) denotes the Frobenius norm.

The weighted MMD and CORAL losses were incorporated into the overall loss function during training. The combined loss is written as:
\begin{equation}
\mathcal{L}_{\text{total}} = \alpha \mathcal{L}_{\text{classification}} + (1 - \alpha) \mathcal{L}_{\text{domain}},
\label{eq:total}
\end{equation}
where \( \mathcal{L}_{\text{classification}} \) is the cross-entropy loss, and \( \mathcal{L}_{\text{domain}} \) is the sum of the weighted domain adaptation losses:
\begin{equation}
\mathcal{L}_{\text{domain}} = \beta \mathcal{L}_{\text{MMD}} + (1 - \beta) \mathcal{L}_{\text{CORAL}}.
\label{eq:domain}
\end{equation}

By incorporating feature-level weights according to the K-S statistics, the model successfully focuses its attention on matching features with higher domain mismatches, thereby enriching the domain adaptation process and improving generalization to the target dataset.

\section{Results}

\subsection{Case Study}

In this research, we utilized two distinct datasets. The source dataset, derived from the literature \cite{taha2021power}, includes samples collected from the Egyptian Electrical Utility \cite{EGHCDGA2016} and the Indian Utility in the TIFAC laboratory \cite{TIFAC}, comprising 343 samples which were then increased to 887 samples when data augmentation methods were applied. The target dataset obtained from the IEC TC 10 database is 140 samples in size. Both datasets have five fault types: Partial Discharge (PD), Low Energy Discharge (D1), High Energy Discharge (D2), Low and Medium Thermal Fault (T1\&T2), and High Thermal Fault (T3).

\begin{figure}[tb]
\centering
    \includegraphics[width=0.95\columnwidth]{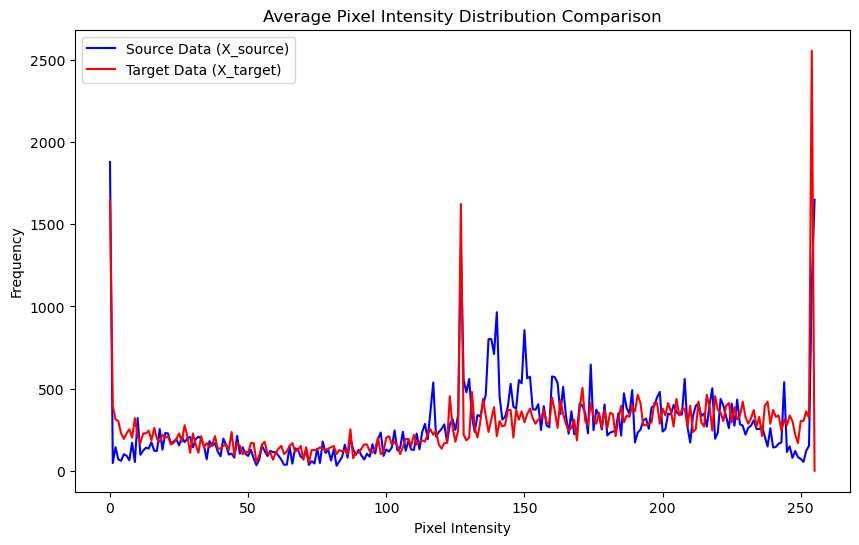}
\caption{Average Pixel Intensity Distribution Comparison for GAF Images of Source and Target Datasets.}
\label{fig_pixel_intensity}
\end{figure}

To quantify the level of difference between the source and target datasets, two statistical measures were employed. At the feature level, the Average Kullback-Leibler Divergence (AKLD) \cite{lin2022transfer} was 0.698, which represents differences in individual feature distributions. In addition, the average pixel intensity distributions of the GAF images, shown in Fig.~\ref{fig_pixel_intensity}, show that the target dataset has a more balanced intensity distribution between mid-range values, whereas the source dataset is more variable, which represents structural differences between domains. These differences also refer to the distributional shift between data sets. These differences point towards the significance of domain adaptation since model transfer may be affected by non-aligned source and target distributions.

The K-S statistic results for individual features can be seen in Table~\ref{tab:ks_results}. 

\begin{table}[tb]
\centering
\caption{K-S Statistics for Features}
\label{tab:ks_results}
\begin{tabular}{ll ll}
\hline
\textbf{Feature} & \textbf{K-S Statistic} & \textbf{Feature} & \textbf{K-S Statistic} \\
\hline
H\textsubscript{2} Ratio & 0.1593 & C\textsubscript{2}H\textsubscript{6} Ratio & 0.0507 \\
CH\textsubscript{4} Ratio & 0.1163 & C\textsubscript{2}H\textsubscript{4} Ratio & 0.1653 \\
C\textsubscript{2}H\textsubscript{2} Ratio & 0.1022 & Ln(C\textsubscript{2}H\textsubscript{6}/CH\textsubscript{4}) & 0.0418 \\
Ln(C\textsubscript{2}H\textsubscript{4}/C\textsubscript{2}H\textsubscript{6}) & 0.1311 & Ln(C\textsubscript{2}H\textsubscript{2}/C\textsubscript{2}H\textsubscript{4}) & 0.0564 \\
Ln(CH\textsubscript{4}/H\textsubscript{2}) & 0.0665 &  &  \\
\hline
\end{tabular}
\end{table}

\subsection{Fault Detection Results}

In order to evaluate the performance of the proposed feature-weighted domain adaptation technique for power transformer fault detection, we conducted extensive experiments among three various domain adaptation techniques: the proposed method (\textit{i.e.,} MCW), MC \cite{liu2022deep}, and Fine-Tuning \cite{yosinski2014transferable}. Fine-Tuning is the method where the model is trained on the source data and then the weights of the model are updated using target dataset samples.
In this paper, two commonly used evaluation metrics, accuracy (ACC) and F1-score, are used.

\begin{figure}[tb]
\centering
    \includegraphics[width=0.95\columnwidth]{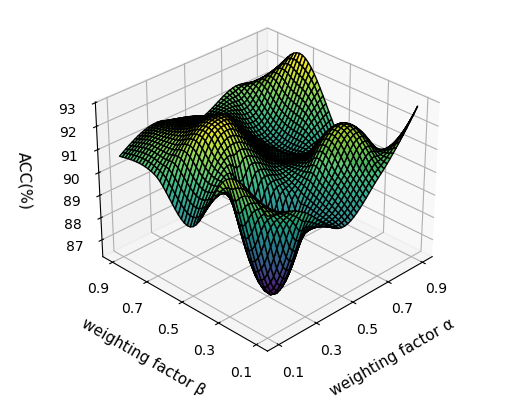}
\caption{Accuracy (\%) comparison for different values of $\alpha$ and $\beta$.}
\label{fig_alpha_beta}
\end{figure}

The effect of critical hyperparameters and model configurations on classification accuracy was also examined. Our domain adaptation model is regulated by two weighting parameters, $\alpha$ and $\beta$, which control the relative trade-off between classification and domain alignment losses \eqref{eq:total}, \eqref{eq:domain}. To find optimal settings, several configurations were tested, as illustrated in Fig.~\ref{fig_alpha_beta}. Optimal classification accuracy was obtained when $\alpha = 0.5$ and $\beta = 0.7$, and these settings were used for further experiments.

\begin{figure}[tb]
\centering
    \includegraphics[width=0.95\columnwidth]{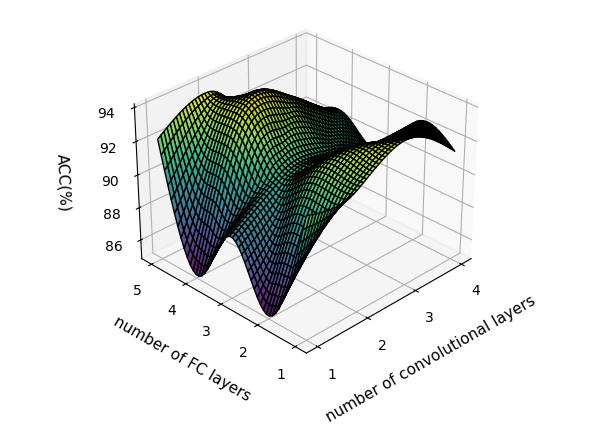}
\caption{Accuracy (\%) comparison for different model architectures.}
\label{fig_num_layers}
\end{figure}

Similarly, the impact of varying the number of convolutional and FC layers was examined. As shown in Fig.~\ref{fig_num_layers}, different configurations resulted in different levels of accuracy. The configuration comprising two convolutional layers and four FC layers was selected as it achieved high accuracy while maintaining a relatively low model complexity. This structure was, therefore, adopted for all subsequent experiments.

The proposed feature-weighted domain adaptation method (MCW) was evaluated in comparison with MC and Fine-Tuning across five transformer fault categories. Table~\ref{tab:running_state} presents the accuracy and F1-score for each method.

\begin{table}[tb]
\centering
\small
\caption{Fault Detection Performance of Different Methods}
\renewcommand{\arraystretch}{1.2}
\setlength{\tabcolsep}{3pt}
\begin{tabular}{ccccccc}
\hline
\multirow{2}{*}{\textbf{Fault Type}} & \multicolumn{3}{c}{\textbf{Accuracy (\%)}} & \multicolumn{3}{c}{\textbf{F1-score (\%)}} \\
\cline{2-7}
 & \textbf{Fine-Tuning} & \textbf{MC} & \textbf{MCW} & \textbf{Fine-Tuning} & \textbf{MC} & \textbf{MCW} \\
\hline
D1  & 78.6  & 89.3  & 85.7  & 81.5  & 89.3  & 90.6  \\
D2  & 82.1  & 82.1  & 96.4  & 82.1  & 88.5  & 94.7  \\
PD  & 92.9  & 96.4  & 100.0  & 89.7  & 93.1  & 96.6  \\
T1\&T2  & 82.1  & 100.0  & 89.3  & 83.6  & 91.8  & 92.6  \\
T3  & 92.9  & 89.3  & 96.4  & 91.2  & 94.3  & 93.1  \\
\hline
\textbf{Average} & 85.7  & 91.4  & \textbf{93.6}  & 85.6  & 91.4  & \textbf{93.5}  \\
\hline
\end{tabular}
\label{tab:running_state}
\end{table}

The results demonstrate that the MCW method outperforms MC and Fine-Tuning techniques across multiple runs, achieving 2.2\% and 7.9\% higher accuracy, respectively, and 2.1\% and 7.9\% greater F1-scores, on average. These improvements underscore the effectiveness of incorporating feature weighting in domain adaptation, leading to more precise alignment between source and target distributions.

To further assess the robustness of each domain adaptation method under varying target sample sizes, we evaluated classification performance across different proportions of the target dataset. As shown in Table~\ref{tab:accuracy_comparison}, the accuracy increases when more target samples are used in training. Notably, even when trained with only 30\% of the target dataset, our method achieved an accuracy of 85.9\%, demonstrating its capability to perform well even with limited labeled target data. Furthermore, the MCW method consistently outperforms MC and Fine-Tuning across all dataset sizes. This consistent improvement highlights the advantage of feature-weighted domain adaptation in mitigating distributional discrepancies between source and target datasets.

\begin{table}[tb]
\centering
\caption{Accuracy (\%) for Different Methods Across Various Sample Sizes}
\renewcommand{\arraystretch}{1.3}
\setlength{\tabcolsep}{6pt}
\begin{tabular}{cccc}
\hline
\textbf{Training Samples} & \textbf{Fine-Tuning} & \textbf{MC} & \textbf{MCW} \\
\hline
30\%  & 82.1  & 85.0  & \textbf{85.9}  \\
50\%  & 83.6  & 87.1  & \textbf{88.1}  \\
70\%  & 85.7  & 91.4  & \textbf{93.6}  \\
90\%  & 86.4  & 92.1  & \textbf{95.3}  \\
\hline
\end{tabular}
\label{tab:accuracy_comparison}
\end{table}

\section{Conclusions}
This study proposed a feature-weighted domain adaptation method for power transformer fault diagnosis, integrating MMD-CORAL with feature-specific weighting. By employing the K-S statistic, the model prioritized features with greater distributional differences, enhancing domain adaptation effectiveness. Experimental results demonstrated that the proposed method consistently outperformed MC and Fine-Tuning approaches. Specifically, the MCW method achieved up to 7.9\% higher accuracy and 7.9\% greater F1-score compared to Fine-Tuning, and 2.2\% and 2.1\% improvements over the MC method, respectively. Remarkably, the model maintained high performance even with limited target samples, achieving 85.9\% accuracy when trained with only 30\% of the target dataset, demonstrating its robustness in data-scarce scenarios. The key advantage of the proposed method lies in its adaptive feature weighting strategy, which ensures that domain adaptation efforts focus on aligning the most influential features, thereby improving model generalization. Additionally, the model demonstrated higher stability across varying target sample sizes, confirming its adaptability to diverse real-world data conditions. These findings highlight the importance of feature weighting in domain adaptation for power transformer fault diagnosis. Future work can explore integrating the proposed model with additional domain adaptation methods using ensemble learning and extending its application to real-time monitoring systems.

\section*{Acknowledgment}
We would like to express our gratitude for the financial support provided
by the Natural Sciences and Engineering Research Council of Canada (NSERC), [NSERC Discovery Grant No. RGPIN-2023-05408].

\end{document}